# Linear Attention Mechanism: An Efficient Attention for Semantic Segmentation

Rui Li*, Jianlin Su*, Chenxi Duan, and Shunyi Zheng

*Abstract*—The attention mechanism can meticulously refine the feature maps and eminently boost the performance of the network, and has become an essential technology applied in computer vision and natural language processing. However, the memory and computational costs of the dot-product attention mechanism widely used in vision and language tasks increase quadratically with the spatio-temporal size of the input. Such growth hinders the usage of attention mechanism in application scenarios with large inputs, e.g., large-scale videos, long sequences, or high-resolution images. In this paper, to remedy this deficiency, we propose a Linear Attention Mechanism which is approximate to dot-product attention with much less memory and computational costs. The efficient design makes the incorporation between attention mechanisms and neural networks more flexible and versatile. Experiments conducted on semantic segmentation demonstrated the effectiveness of linear attention mechanism. Code is available at https://github.com/lironui/Linear-Attention-Mechanism.

## I. Introduction

Attention mechanisms, bolstered by their powerful capability of exploiting long-range dependencies of feature maps and facilitating neural networks to attain global contextual information, are at the forefront of convolution and recurrence researches. Dot-product attention mechanism, which generates the response at every pixel by weighting features at all pixels in the previous layer, expands the receptive field to the whole input feature maps in one pass. Benefitting from their specialty on capturing long-range dependencies, dot-product attention mechanisms have been widely used in vision and language tasks. Dot-product-attention-based Transformer has manifested the state-of-the-art performances in nearly all tasks in natural language processing. The non-local module [1], a dot-product-based attention modified for computer vision, has demonstrated its superiority on image classification [2], object detection [3], semantic segmentation [4], and panoptic segmentation [5].

However, as the memory and computational costs of the dot-product attention mechanism increase quadratically with the spatio-temporal size of the input, how to model global dependency on large inputs, e.g. large-scale videos, long sequences, or high-resolution images, remains an intractable problem. To alleviate the enormous computational consumptions, Child [6] designed a sparse factorizations of the attention matrix and reduce the complexity from $O(N^2)$ to $O(N\sqrt{N})$. Using locality sensitive hashing, Kitaev [7] reduced the complexity to $O(N \log N)$. To further reduce the complexity to $O(N)$, Katharopoulos [8] taken self-attention as a linear dot-product of kernel feature maps, and Shen [9] modified the position of softmax functions.

In this paper, we reduce the complexity of dot-product attention mechanism to $O(N)$ from another facet. And experiments conducted on semantic segmentation manifest the effectiveness of proposed linear attention mechanism. The major contribution of this paper could be listed as follows:
1) We proposed a linear attention mechanism which reduce the complexity of attention mechanism from $O(N^2)$ to $O(N)$.
2) The linear attention mechanism allows the combination between attention modules and neural networks more flexible and versatile.
3) Performances of several baseline networks are boosted by the linear attention mechanism on semantic segmentation.

## II. Related works

### A. Dot-Product Attention

To enhance word alignment in machine translation, Bahdanau [10] proposed the initial formulation of the dot-product attention mechanism. Successively, in Transformer [11], recurrences are completely replaced by attention. And the state-of-the-art records in almost all tasks in natural language processing tasks demonstrate the superiority of attention mechanism. Wang [1] modified the dot-product attention for computer vision and proposed the non-local module. Subsequent works conducted on many tasks of computer vison including image classification [2], object detection [3], semantic segmentation [4], and panoptic segmentation [5] further prove the effectiveness and universality of attention mechanism. Meanwhile, attention is an effective technology for speech recognition [12, 13].

*These authors contributed equally to this work.
This work was supported in part by the National Natural Science Foundations of China (No. 41671452). *(Corresponding author: Rui Li.)*
R. Li and S. Zheng are with School of Remote Sensing and Information Engineering, Wuhan University, Wuhan 430079, China (e-mail: lironui@whu.edu.cn; syzheng@whu.edu.cn).
Jianlin Su is with the Technology Co., Ltd. (e-mail: bojonesu@wezhuiyi.com)
C. Duan is with the State Key Laboratory of Information Engineering in Surveying, Mapping, and Remote Sensing, Wuhan University, Wuhan 430079, China; chenxiduan@whu.edu.cn (e-mail: chenxiduan@whu.edu.cn).

We design linear attention mechanism on top of Dual attention network (DANet) [4], and evaluate the performance of linear attention on semantic segmentation.

### B. Scaling Attention

Besides dot-product attention, there is another genre of techniques referred as attention in the literatures. To distinguish them, this section calls them as scaling attention. Unlike dot-product attention which models global dependency, scaling attention reinforces the informative features and whittles the information-lacking features. In the squeeze-and-excitation (SE) module [14], a global average pooling layer and a linear layer are harnessed to calculated a scaling factor for each channel and then weights the channels accordingly. By adding global max pooling layer beside global average pooling and a spatial attention submodule, convolutional block attention module (CBAM) [15] further boost the performance of SE block.

Actually, whether the principles or purposes of dot-product attention and scaling attention are completely divergent. In this paper, we are going to focus on dot-product attention.

### C. Semantic Segmentation

Fully Convolutional Network (FCN) based methods have brought huge progress in semantic segmentation. DilatedFCN and EncoderDecoder are two prominent directions followed by FCN. In DilatedFCNs [16-22], dilate or atrous convolutions are harnessed to retain the receptive field of view and a multi-scale context module id utilized to cope with high-level feature maps. Alternatively, EncoderDecoders [23-27] utilize an encoder to capture multi-level feature maps, which are then incorporated into the final prediction by a decoder.

**DilatedFCN** The dilated or atrous convolution [16, 17] has been proven to be an effectual technology for dense prediction and has been successfully utilized in semantic segmentation. In DeepLab [18, 19], atrous spatial pyramid pooling (ASPP) which comprises parallel dilated convolutions with diverse dilated rates could embed contextual information, while a pyramid pooling module (PPM) is included in PSPNet [20] to collect the effective contextual prior between different scales. Alternatively, EncNet [21] proposes a context encoding module to exploit global contextual information. Then FastFCN [22] replace dilated convolutions to a joint pyramid up sampling module (JPU) which could reduce computation complexity.

**EncoderDecoder** Skip connections, which combine the high-level features generated by decoder and low-level features generated by corresponding encoder, are proposed to construct U-Net [23]. In U-Net++ [24], nested and dense skip connections substitute the plain skip connections in U-Net, which narrow the semantic gap between the encoder and decoder. U-Net 3+ [25] and MACU-Net [26] further propose full-scale skip connections and multi-scale skip connections to enhance the capability of skip connections. Meanwhile, RefineNet [27] design a multipath refinement structure, which captures all the available information along the down-sampling process. Taking DeepLab V3 as encoder, DeepLab V3+ [19] combines the merits of DilatedFCN and EncoderDecoder.

## III. METHODOLOGY

In this section, we analysis the dot-product attention and formalize proposed linear attention mechanism. We illustrate that substituting the attention from the conventional softmax attention to first-order approximation of Taylor expansion lead to linear time and memory complexity.

### A. Definition of Dot-Product Attention

Supposing $N$ and $D_x$ denote the length of input sequences and the number of input dimensions, given a feature $\boldsymbol{X} = [\boldsymbol{x}_1, \cdots, \boldsymbol{x}_N] \in \mathbb{R}^{N \times D_x}$, dot-product attention utilize three projected matrices $\boldsymbol{W}_q \in \mathbb{R}^{D_x \times D_k}$, $\boldsymbol{W}_k \in \mathbb{R}^{D_x \times D_k}$, and $\boldsymbol{W}_v \in \mathbb{R}^{D_x \times D_v}$ to generate corresponding query matrix $\boldsymbol{Q}$, the key matrix $\boldsymbol{K}$, and the value matrix $\boldsymbol{V}$:

$$\boldsymbol{Q} = \boldsymbol{X}\boldsymbol{W}_q \in \mathbb{R}^{N \times D_k},$$
$$\boldsymbol{K} = \boldsymbol{X}\boldsymbol{W}_k \in \mathbb{R}^{N \times D_k}, \qquad (1)$$
$$\boldsymbol{V} = \boldsymbol{X}\boldsymbol{W}_v \in \mathbb{R}^{N \times D_v}.$$

The dimensions of query matrix and key matrix must be identical. Then a normalization function $\rho$ evaluates the similarity between the $i$-th query feature $\boldsymbol{q}_i^T \in \mathbb{R}^{D_k}$ and the $j$-th key feature $\boldsymbol{k}_j \in \mathbb{R}^{D_k}$ by $\rho(\boldsymbol{q}_i^T \boldsymbol{k}_j) \in \mathbb{R}^1$. Generally, as the query feature and key feature are generated by diverse layers, the similarities between $\rho(\boldsymbol{q}_i^T \boldsymbol{k}_j)$ and $\rho(\boldsymbol{q}_j^T \boldsymbol{k}_i)$ are not symmetric. By calculating the similarities between all pairs of positions and taking the similarities as weights, the dot-product attention module computes the value at position $i$ via aggregating the value features from all positions based on weighted summation:

$$D(\boldsymbol{Q}, \boldsymbol{K}, \boldsymbol{V}) = \rho(\boldsymbol{Q}\boldsymbol{K}^T)\boldsymbol{V}. \qquad (2)$$

The softmax is the common normalization function:

$$\rho(\boldsymbol{Q}^T \boldsymbol{K}) = softmax_{row}(\boldsymbol{Q}\boldsymbol{K}^T), \qquad (3)$$

where $softmax_{row}$ indicates applying the softmax function along each row of matrix $\boldsymbol{Q}\boldsymbol{K}^T$.

The $\rho(\boldsymbol{Q}\boldsymbol{K}^T)$ models the similarities between all pairs of positions. However, as $\boldsymbol{Q} \in \mathbb{R}^{N \times D_k}$ and $\boldsymbol{K}^T \in \mathbb{R}^{D_k \times N}$, product between $\boldsymbol{Q}$ and $\boldsymbol{K}^T$ belongs to $\mathbb{R}^{N \times N}$, which leads to $O(N^2)$ memory complexity and $O(N^2)$ computational complexity. Therefore, the high resource demanding of dot-product greatly limit the application on large inputs. An improvement method is modifying the softmax [9], and another is introducing kernel method.

### B. Generalization of Dot-Product Attention Based on Kernel

Under condition of softmax normalization function, the $i$-th row of result matrix generated by dot-product attention module according to equation (2) can be written as:

$$D(\boldsymbol{Q}, \boldsymbol{K}, \boldsymbol{V})_i = \frac{\sum_{j=1}^{N} e^{\boldsymbol{q}_i^T \boldsymbol{k}_j} \boldsymbol{v}_j}{\sum_{j=1}^{N} e^{\boldsymbol{q}_i^T \boldsymbol{k}_j}}, \qquad (4)$$

Then, equation (4) can be generalized for any normalization function and rewritten as:

$$D(\boldsymbol{Q}, \boldsymbol{K}, \boldsymbol{V})_i = \frac{\sum_{j=1}^{N} sim(\boldsymbol{q}_i, \boldsymbol{k}_j)\boldsymbol{v}_j}{\sum_{j=1}^{N} sim(\boldsymbol{q}_i, \boldsymbol{k}_j)}, \qquad (5)$$
$$sim(\boldsymbol{q}_i, \boldsymbol{k}_j) \geq 0.$$

If $sim(\boldsymbol{q}_i, \boldsymbol{k}_j) = e^{\boldsymbol{q}_i^T \boldsymbol{k}_j}$, equation (5) is equivalent to equation (4). And $sim(\boldsymbol{q}_i, \boldsymbol{k}_j)$ can be further expanded as $sim(\boldsymbol{q}_i, \boldsymbol{k}_j) = \phi(\boldsymbol{q}_i)^T \varphi(\boldsymbol{k}_j)$, where $\phi(\cdot)$ and $\varphi(\cdot)$ can be seen as kernel smoothers [28]. Then equation (4) can be rewritten as:

$$D(\boldsymbol{Q}, \boldsymbol{K}, \boldsymbol{V})_i = \frac{\sum_{j=1}^{N} \phi(\boldsymbol{q}_i)^T \varphi(\boldsymbol{k}_j) \boldsymbol{v}_j}{\sum_{j=1}^{N} \phi(\boldsymbol{q}_i)^T \varphi(\boldsymbol{k}_j)}, \quad (6)$$

and then can simplified as:

$$D(\boldsymbol{Q}, \boldsymbol{K}, \boldsymbol{V})_i = \frac{\phi(\boldsymbol{q}_i)^T \sum_{j=1}^{N} \varphi(\boldsymbol{k}_j) \boldsymbol{v}_j^T}{\phi(\boldsymbol{q}_i)^T \sum_{j=1}^{N} \varphi(\boldsymbol{k}_j)}. \quad (7)$$

As $\boldsymbol{K} \in \mathbb{R}^{D_k \times N}$ and $\boldsymbol{V}^T \in \mathbb{R}^{N \times D_v}$, product between $\boldsymbol{K}$ and $\boldsymbol{V}^T$ belongs to $\mathbb{R}^{D_k \times D_v}$, which considerably reduces the complexity of dot-product attention mechanism. For example, [8] adopts $\phi(x) = \varphi(x) = \text{elu}(x) + 1$.

### C. Linear Attention Mechanism

Different previous researches, we conceive linear attention mechanism based on first-order approximation of Taylor expansion on equation (4):

$$e^{\boldsymbol{q}_i^T \boldsymbol{k}_j} \approx 1 + \boldsymbol{q}_i^T \boldsymbol{k}_j. \quad (8)$$

However, the above approximation cannot guarantee the non-negativity. To ensure $\boldsymbol{q}_i^T \boldsymbol{k}_j \geq -1$, we can normalize $\boldsymbol{q}_i$ and $\boldsymbol{k}_j$ by $l_2$ norm:

$$sim(\boldsymbol{q}_i, \boldsymbol{k}_j) = 1 + \left(\frac{\boldsymbol{q}_i}{\|\boldsymbol{q}_i\|_2}\right)^T \left(\frac{\boldsymbol{k}_j}{\|\boldsymbol{k}_j\|_2}\right). \quad (9)$$

Then, equation (5) can be rewritten as:

$$D(\boldsymbol{Q}, \boldsymbol{K}, \boldsymbol{V})_i = \frac{\sum_{j=1}^{N} \left(1 + \left(\frac{\boldsymbol{q}_i}{\|\boldsymbol{q}_i\|_2}\right)^T \left(\frac{\boldsymbol{k}_j}{\|\boldsymbol{k}_j\|_2}\right)\right) \boldsymbol{v}_j}{\sum_{j=1}^{N} \left(1 + \left(\frac{\boldsymbol{q}_i}{\|\boldsymbol{q}_i\|_2}\right)^T \left(\frac{\boldsymbol{k}_j}{\|\boldsymbol{k}_j\|_2}\right)\right)}, \quad (10)$$

and simplified as:

$$D(\boldsymbol{Q}, \boldsymbol{K}, \boldsymbol{V})_i = \frac{\sum_{j=1}^{N} \boldsymbol{v}_j + \left(\frac{\boldsymbol{q}_i}{\|\boldsymbol{q}_i\|_2}\right)^T \sum_{j=1}^{N} \left(\frac{\boldsymbol{k}_j}{\|\boldsymbol{k}_j\|_2}\right) \boldsymbol{v}_j^T}{N + \left(\frac{\boldsymbol{q}_i}{\|\boldsymbol{q}_i\|_2}\right)^T \sum_{j=1}^{N} \left(\frac{\boldsymbol{k}_j}{\|\boldsymbol{k}_j\|_2}\right)}. \quad (11)$$

The above equation can be written in vectorized form as:

$$D(\boldsymbol{Q}, \boldsymbol{K}, \boldsymbol{V}) = \frac{\sum_j \boldsymbol{V}_{i,j} + \left(\frac{\boldsymbol{Q}}{\|\boldsymbol{Q}\|_2}\right) \left(\left(\frac{\boldsymbol{K}}{\|\boldsymbol{K}\|_2}\right)^T \boldsymbol{V}\right)}{N + \left(\frac{\boldsymbol{Q}}{\|\boldsymbol{Q}\|_2}\right) \sum_j \left(\frac{\boldsymbol{K}}{\|\boldsymbol{K}\|_2}\right)_{i,j}^T}. \quad (12)$$

As $\sum_{j=1}^{N} \left(\frac{\boldsymbol{k}_j}{\|\boldsymbol{k}_j\|_2}\right) \boldsymbol{v}_j^T$ and $\sum_{j=1}^{N} \left(\frac{\boldsymbol{k}_j}{\|\boldsymbol{k}_j\|_2}\right)$ can be calculated and reused for every query, time and memory complexity of the proposed linear attention mechanism based on equation (12) is $O(N)$.

## IV. EXPERIMENTAL RESULTS

### A. Dataset

The effectiveness of linear attention mechanism is verified using fine Gaofen Image Dataset (GID) [29]. Fine GID contains 10 RGB images in the size of $7200 \times 6800$ captured by Gaofen 2 Satellite in China. Each image covering a geographic region of 506 $km^2$. The images contained in fine GID are labeled with fifteen classes. We separately partition each image into non-overlap patch sets with the size of $256 \times 256$, and just discard the pixels on the edges which cannot be divisible by 256. Thus, 7280 patches are obtained. Then we randomly select 60% patches as training set, 20% patches as validation set, and the rest 20% patches as test set.

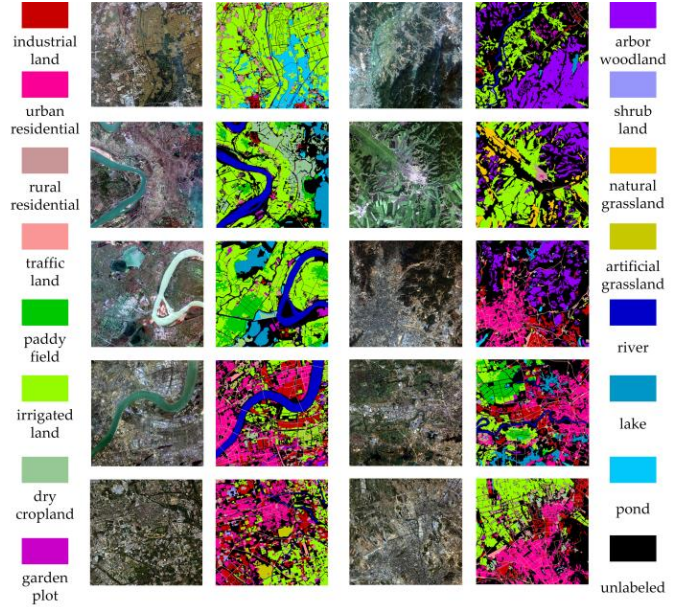

Fig. 5. Fine Gaofen Image Dataset.

TABLE I
THE SAMPLES FOR EACH CATEGORY FOR TRAINING, VALIDATION AND TEST.

| Class | Train | Val | Test |
| --- | --- | --- | --- |
| industrial land | 12590501 | 4826071 | 4006917 |
| urban residential | 24896380 | 7946414 | 8614169 |
| rural residential | 9815607 | 3814329 | 3348257 |
| traffic land | 8517348 | 2905318 | 2812296 |
| paddy field | 7412904 | 2956311 | 2885346 |
| irrigated land | 66134918 | 22212423 | 21407788 |
| dry cropland | 4933351 | 1862748 | 2249710 |
| garden plot | 1540615 | 507740 | 415271 |
| arbor woodland | 16043641 | 5632430 | 5324376 |
| shrub land | 485079 | 205920 | 164203 |
| natural grassland | 2696185 | 1057328 | 1253542 |
| artificial grassland | 1737725 | 440031 | 402009 |
| river | 10812394 | 3089753 | 3401676 |
| lake | 5667907 | 1815322 | 2008129 |
| pond | 5524311 | 1708548 | 1784859 |

### B. Experimental Setting

All of the models are implemented with PyTorch, and the optimizer is set as Adam with 0.0003 learning rate and 16 batch size. All the experiments are implemented on a single NVIDIA GeForce RTX 2080ti GPU with 11 GB RAM. The cross-entropy loss function is used as quantitative evaluation and backpropagation index to measure the disparity between the obtained 2D segmentation maps and ground truth.

For each dataset, the overall accuracy (OA), average accuracy (AA), Kappa coefficient (K), mean Intersection over Union (mIoU), and F1-score (F1) are adopted as evaluation indexes.

## C. Results on Fine GID

The experimental results of different methods GID are demonstrated in Table II. The proposed linear attention mechanism enhances the performance of 7 baselines. And we are still conduct more experiments to verify the effectiveness of proposed attention.

## V. CONCLUSION

In this paper, we propose a linear attention mechanism which reduces the complexity of dot-product attention mechanism from $O(N^2)$ to $O(N)$. And experiments conducted on semantic segmentation manifest the effectiveness of proposed linear attention mechanism.

TABLE II
THE EXPERIMENTAL RESULTS ON FINE GID DATASET.

| Method | OA | AA | K | mIoU | F1 |
|---|---|---|---|---|---|
| U-Net | 86.378 | 74.532 | 83.357 | 64.516 | 75.532 |
| U-Net LAM | **87.692** | **77.297** | **84.935** | **68.038** | **78.593** |
| Res101 | 89.251 | 80.451 | 86.846 | 72.433 | 82.510 |
| Res101 LAM | **90.178** | **82.757** | **88.041** | **74.085** | **83.105** |
| RefineNet | 89.857 | 81.169 | 87.597 | 73.167 | 83.113 |
| RefineNet LAM | **90.214** | **83.544** | **88.083** | **74.973** | **84.311** |
| DeepLab | 89.388 | 80.905 | 87.079 | 71.809 | 81.077 |
| DeepLabV3 LAM | **89.576** | **81.692** | **87.287** | **72.827** | **82.702** |
| DeepLabV3+ | 90.125 | 81.483 | 87.959 | 72.668 | 81.492 |
| DeepLabV3+ LAM | **90.315** | **81.695** | **88.182** | **73.727** | **82.736** |
| PSPNet | 90.573 | 82.211 | 88.485 | 74.797 | 83.761 |
| PSPNet LAM | **90.725** | **83.088** | **88.677** | **75.695** | **84.480** |
| FastFCN | 90.336 | 83.625 | 88.221 | 74.364 | 83.704 |
| FastFCN LAM | **90.835** | 83.075 | **88.769** | **75.174** | **84.023** |